\documentclass[lettersize,journal]{IEEEtran}
\usepackage{multicol}
\usepackage{times}
\usepackage{amssymb, amsmath}
\usepackage{graphicx}
\usepackage{float}
\usepackage{booktabs}
\usepackage{epstopdf}
\usepackage[table]{xcolor}
\usepackage{amsmath}
\usepackage{numprint}
\usepackage{subcaption}
\usepackage{caption}
\usepackage{multirow}
\usepackage{adjustbox}
\usepackage{subfig}
\usepackage{xcolor}
\usepackage{array}
\usepackage{calc}
\usepackage{calc}
\usepackage{tikz}
\usetikzlibrary{positioning}
\usepackage{algorithmic}
\usepackage{algorithm}
\usepackage{xcolor}
\usepackage{subcaption}
\PassOptionsToPackage{export}{adjustbox}
\usepackage{adjustbox}
\usepackage{array}
\newcolumntype{M}[1]{>{\centering\arraybackslash}m{#1}}
\newcolumntype{N}{@{}m{0pt}@{}}

\definecolor{antiquefuchsia}{rgb}{0.57, 0.36, 0.51}
\definecolor{ao(english)}{rgb}{0.0, 0.5, 0.0}
\definecolor{blue(ncs)}{rgb}{0.0, 0.53, 0.74}
\definecolor{byzantium}{rgb}{0.44, 0.16, 0.39}
\definecolor{copper}{rgb}{0.72, 0.45, 0.2}
\definecolor{ballblue}{rgb}{0.13, 0.67, 0.8}
\definecolor{blue(munsell)}{rgb}{0.0, 0.5, 0.69}
\definecolor{cordovan}{rgb}{0.54, 0.25, 0.27}
\definecolor{darkmagenta}{rgb}{0.55, 0.0, 0.55}

\usepackage{subcaption}

\usepackage[numbers,square, sort&compress]{natbib}

\ifCLASSINFOpdf

\else

\fi

\usepackage{filecontents,lipsum}

\begin{document}
\title{LLM-as-Judge for Semantic Judging of Powerline Segmentation in UAV Inspection}
\author{Akram Hossain$^{1}$,
Rabab Abdelfattah$^{1}$~\IEEEmembership{Senior Member, IEEE}, \\ Xiaofeng Wang$^{2}$~\IEEEmembership{Member, IEEE}, 
Kareem Abdelfatah$^{3}$~\IEEEmembership{Member, IEEE} \\
\thanks{Corresponding author: Rabab Abdelfattah (e-mail: rabab.abdelfattah@usm.edu)}
\thanks{A. Hossain, and R. Abdelfattah are with the School of Computing Sciences and Computer Engineering, The University of Southern Mississippi, Hattiesburg, MS 39406 USA.}
\thanks{Xiaofeng Wang is with the Electrical Engineering Department, University of South Carolina, Columbia, SC, USA.}
\thanks{Kareem Abdelfatah is with the Computer Science Department, Faculty of Computers \& Artificial Intelligence, Fayoum University, Fayoum, Egypt.}
}

\maketitle

\begin{abstract}
The deployment of lightweight segmentation
models on drones for autonomous power line inspection
presents a critical challenge: maintaining reliable performance
under real-world conditions that differ from training data.
Although compact architectures such as U-Net enable real-time
onboard inference, their segmentation outputs can degrade
unpredictably in adverse environments, raising safety concerns.
In this work, we study the feasibility of using a large language
model (LLM) as a semantic judge to assess the reliability of
power line segmentation results produced by drone-mounted
models. Rather than introducing a new inspection system, we
formalize a watchdog scenario in which an offboard LLM
evaluates segmentation overlays and examine whether such a
judge can be trusted to behave consistently and perceptually
coherently. To this end, we design two evaluation protocols that
analyze the judge’s repeatability and sensitivity. First, we assess repeatability by repeatedly querying the LLM with identical
inputs and fixed prompts, measuring the stability of its quality
scores and confidence estimates. Second, we evaluate perceptual
sensitivity by introducing controlled visual corruptions (fog,
rain, snow, shadow, and sunflare) and analyzing how the judge’s
outputs respond to progressive degradation in segmentation
quality. Our results show that the LLM produces highly
consistent categorical judgments under identical conditions
while exhibiting appropriate declines in confidence as visual
reliability deteriorates. Moreover, the judge remains responsive
to perceptual cues such as missing or misidentified power
lines, even under challenging conditions. These findings suggest
that, when carefully constrained, an LLM can serve as a
reliable semantic judge for monitoring segmentation quality in
safety-critical aerial inspection tasks.

\end{abstract}


\section{Introduction}

The automation of powerline inspection using unmanned aerial vehicles represents a major step toward scalable and cost-effective infrastructure maintenance. Modern deep learning models have demonstrated strong performance in segmenting thin structures such as powerlines under controlled benchmark conditions \cite{foundation2026, abdelfattah2023plgan, hossain2025evaluating}. However, once deployed in real-world environments, these models are exposed to a wide range of previously unseen conditions, including adverse weather, changing illumination, motion blur, partial occlusions, and sensor noise. Under such conditions, segmentation quality may degrade silently, without any ground truth available to signal failure. This gap between offline evaluation and online deployment poses a fundamental challenge for safety-critical inspection systems.

Traditional performance metrics such as intersection over union or pixel accuracy are inherently tied to the availability of ground truth annotations. As a result, they become unusable once a model is deployed in the field. In practice, this leaves autonomous inspection systems without a principled mechanism to assess whether their visual perception remains reliable. The current safeguard relies on human operators who manually review video streams or flagged frames. While effective in limited settings, human-in-the-loop monitoring does not scale to continuous, long-duration, or large-fleet deployments and undermines the autonomy promised by aerial inspection platforms. This work addresses a challenging practical question: how can the performance of a deployed vision model be monitored in real time in the absence of ground truth?

Instead of estimating conventional accuracy metrics post-deployment, we propose reframing the problem as one of \emph{semantic safety monitoring}. Specifically, the system should reason about whether the visual output is plausible, structurally consistent, and reliable enough to support downstream decisions. Recent advances in multimodal large language models have revealed strong visual reasoning capabilities that extend beyond classification or captioning. These models can reason about spatial structure, continuity, and semantic plausibility in complex scenes. This observation motivates a new role for such models, not as primary perception engines, but as independent evaluators of perception quality. In this paper, we introduce the \emph{LLM-as-Judge} paradigm, in which a multimodal language model receives an overlay of a segmentation mask on the corresponding image and produces a structured assessment consisting of a discrete quality score, a confidence estimate, and a textual rationale.
\begin{figure*}[!t]
    \centering
        \includegraphics[width=1\textwidth]{ 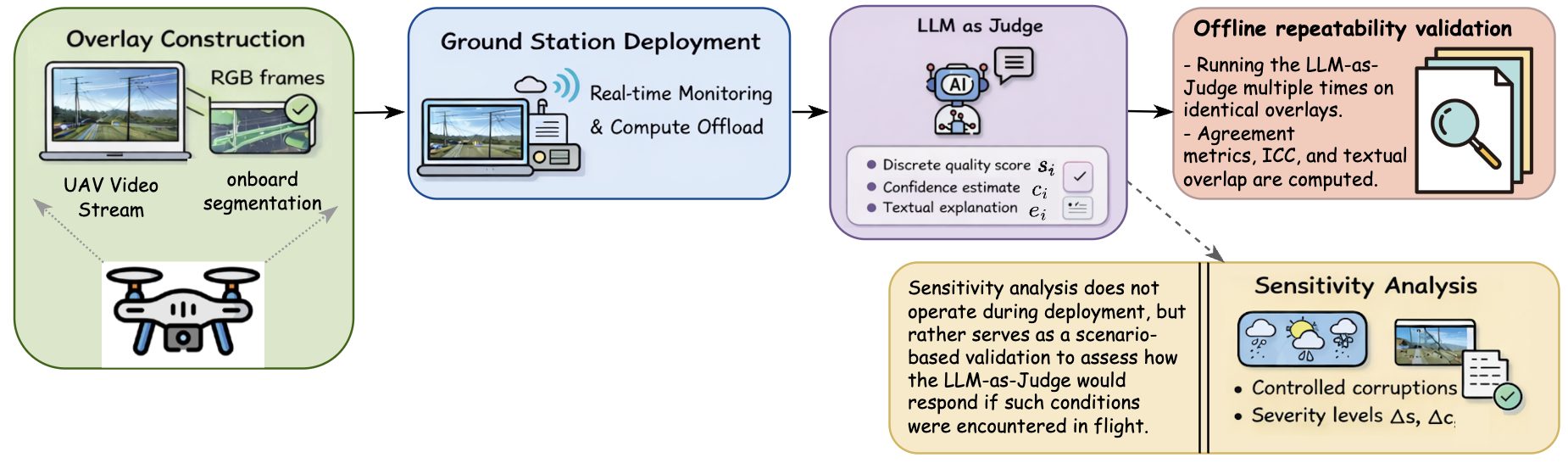}
    \vspace{-3 mm}
    \caption{System overview of the proposed LLM-as-Judge framework for powerline inspection. UAV RGB frames are segmented onboard and streamed as mask--image overlays to a ground station for compute offload. A multimodal LLM produces a discrete quality score $s_i$, confidence estimate $c_i$, and textual explanation $e_i$ for real-time safety monitoring. Offline repeatability and scenario-based sensitivity analyses evaluate output stability and robustness under controlled corruptions.}
    \label{fig:framework}
\end{figure*}
While the LLM-as-Judge paradigm has been widely explored for text-based evaluation and, more recently, for image-level semantic judgments such as safety classification or preference ranking, its application to \emph{thin-object segmentation outputs} has not been systematically studied. In powerline inspection, segmentation errors manifest as broken continuity, missing line segments, or spurious detections—failure modes that are subtle, geometric, and safety-critical, and fundamentally different from semantic misclassification or caption quality. Moreover, once deployed, powerline inspection systems operate without ground truth, making it unclear whether existing LLM-as-Judge methodologies and evaluation criteria are suitable for this setting.

To enable real-time deployment, we design the system such that the segmentation model operates onboard the drone, while the LLM-based judge runs on a ground control station. Video frames and predicted masks are streamed to the ground station, where the judge performs semantic inspection without imposing computational burden on the aerial platform. This architecture supports real-time semantic monitoring and post-mission auditing, with the potential to inform live intervention. The judge can flag critical failure modes such as broken line continuity, spurious detections, or implausible structures, thereby acting as a semantic safety monitor rather than a replacement for the segmentation model. Fig.~\ref{fig:framework} illustrates the overall architecture of the proposed LLM-as-Judge framework, highlighting the separation between onboard perception and ground-based semantic monitoring.

A key challenge in adopting such a system lies in establishing trust in the behavior of the LLM-based judge itself. Language models are known to exhibit stochastic decoding, raising concerns about output stability and consistency under repeated evaluation. Moreover, a judge that produces identical outputs regardless of visual degradation would be perceptually unsuitable for safety monitoring. For this reason, we develop a rigorous evaluation framework that characterizes two essential properties of the LLM-as-Judge: \emph{repeatability} and \emph{sensitivity}. Repeatability measures whether the judge produces stable numeric outputs under identical inputs, while sensitivity evaluates whether its judgments vary coherently in response to controlled degradation of visual evidence. Together, these properties define whether the judge can function as a reliable monitoring component in autonomous inspection pipelines.

By explicitly separating perception from judgment, the proposed framework enables modular integration with existing vision systems and motivates further research into real-time reliability assessment, adaptive autonomy, and explainable safety monitoring. While this work focuses specifically on \emph{powerline (thin-object) segmentation}, the underlying principles may extend to other deployed vision systems where structured outputs must be monitored continuously in the absence of ground truth.

The main contributions of this paper are summarized as follows.
\begin{itemize}
\item We present the first systematic study of \emph{LLM-as-Judge for powerline (thin-object) segmentation}, investigating whether a multimodal language model can reliably assess segmentation quality in safety-critical inspection tasks without ground truth.
\item We reformulate post-deployment evaluation for \emph{powerline inspection} as a semantic safety monitoring problem, shifting the focus from offline accuracy metrics to online reliability assessment under real-world operating conditions.
\item We introduce a rigorous evaluation methodology tailored to powerline segmentation that quantifies the \emph{repeatability} and \emph{perceptual sensitivity} of LLM-based judges using agreement metrics, intraclass correlation analysis, effect sizes, and controlled visual degradation experiments.
\item We demonstrate that, under realistic corruption scenarios affecting powerline imagery, the proposed LLM-as-Judge exhibits stable categorical judgments and conservative confidence behavior, supporting its suitability as a semantic safety monitor for deployed aerial inspection systems.
\end{itemize}

\section{Related Work}
\label{relted}

With the introduction of large language models (LLMs) and their strong reasoning capabilities, researchers have increasingly explored their use as automated evaluators or judges. One of the earliest works to explicitly frame LLMs as evaluators is JudgeLM~\cite{zhu2023judgelm}, which demonstrates that open-source language models can be fine-tuned into scalable judges for open-ended text generation. This work positions LLM-based judges as a practical alternative to both coarse automatic metrics and costly human evaluation.

Despite their effectiveness, LLM judges are known to exhibit systematic biases. Ye et al.~\cite{ye2024justice} formalize this issue by identifying twelve distinct bias types in LLM-based judgment and proposing a framework to measure them quantitatively. While such biases cannot be fully eliminated, subsequent work suggests that LLMs can still serve as effective reliability judges in specific settings such as faithfulness verification and uncertainty assessment, either through direct evaluation or specialized judge training~\cite{li2025generation}.

Beyond bias, the structure of the judging task itself plays a critical role in evaluation quality. Zheng et al.~\cite{zheng2023judging} categorize LLM-based evaluation into three primary paradigms: (i) pairwise comparison, where a judge selects the better of two responses (or declares a tie), (ii) single-answer grading, where a numeric or categorical score is assigned to one output, and (iii) reference-guided grading, where outputs are evaluated against a provided reference. They further demonstrate that judgment accuracy improves when complex evaluation criteria are decomposed into simpler sub-questions. Complementing this perspective, Pan et al.~\cite{pan2024human} advocate for a human-centered evaluation design that emphasizes aspect-level judgments, alignment with expert standards, and transparent explanations from LLM judges.

Nevertheless, recent studies show that LLM-as-a-judge can diverge substantially from subject-matter experts on expert knowledge tasks. Szymanski et al.~\cite{szymanski2025limitations} report that agreement between LLM judges and human experts often remains in the mid-60\% range in domains such as dietetics and mental health, suggesting that LLM judges may overlook domain-critical errors or risks. This gap motivates evaluation pipelines that retain human experts in the loop rather than relying solely on automated judgment. Similarly, Son et al.~\cite{son2023llm} show that while LLMs function well as scalable evaluators with strong cross-lingual transfer and competence in surface-level or comparative assessments, they remain unreliable in identifying factual inaccuracies, cultural errors, and complex reasoning failures, limiting their applicability in high-stakes evaluations.

Several works propose training and architectural strategies to improve the robustness of LLM judges. Saha et al.~\cite{saha2025learning} introduce EvalPlanner, a Thinking-LLM-as-a-Judge framework that explicitly separates evaluation into planning, reasoning, and final judgment stages. By training judges using self-generated evaluation plans and synthetic preference data, their approach improves judgment accuracy and data efficiency across multiple benchmarks. Yu et al.~\cite{yu2025improve} further argue that judging should be treated as a general capability of LLMs rather than a task-specific skill, proposing a two-stage training framework that combines supervised warm-up with preference optimization to enhance both judgment accuracy and general reasoning ability.

Other work focuses on the construction and benchmarking of evaluation datasets. Raju et al.~\cite{raju2024constructing} propose a pipeline for building domain-specific and multilingual evaluation sets tailored to LLM-as-a-judge, noting that common judge benchmarks disproportionately focus on generic English prompts while under-representing specialized domains such as law and medicine. Their method combines manual seeding, embedding-based clustering, and stratified sampling to curate a balanced benchmark that better differentiates strong models and aligns more closely with Chatbot Arena rankings than existing benchmarks.

Despite these advances, LLM judges remain vulnerable to exploitation. Zhao et al.~\cite{zhao2025one} demonstrate that LLM-based judges can be systematically misled by adversarially crafted or stylistically manipulated outputs, revealing that high judge scores do not always correspond to true task quality. In parallel, Schroeder et al.~\cite{schroeder2024can} examine judge reliability by framing it in terms of internal consistency rather than agreement with humans. They show that single-shot or deterministic evaluations often mask substantial judgment variability due to stochastic sampling and introduce a reliability framework based on McDonald’s omega. Wei et al.~\cite{wei2024systematic} similarly study LLM-as-a-judge for alignment evaluation, showing that prompt templates, internal decision inconsistency, and positional and length biases strongly affect reliability, leading to only moderate alignment with human preferences.

While the majority of prior work focuses on text-based inputs, comparatively little attention has been paid to LLM-as-a-judge in vision or multimodal settings. Wang et al.~\cite{wang2025mllm} introduce CLUE, a zero-shot MLLM-as-a-judge framework for image safety that eliminates the need for human-labeled data by decomposing safety constitutions into objectified rules and precondition chains. Their approach combines debiased token-level judgments with selective chain-of-thought reasoning and demonstrates improved reliability and generalization over naive prompting and fine-tuned classifiers.

Chen et al.~\cite{chen2024mllm} present MLLM-as-a-Judge, a large-scale vision--language benchmark that evaluates whether multimodal LLMs can function as judges across scoring, pairwise comparison, and ranking tasks. Their results show that while advanced models such as GPT-4V align reasonably well with human preferences in pairwise judgments, they suffer from significant bias, hallucination, and inconsistency in absolute scoring and ranking scenarios. Finally, Narayanan et al.~\cite{narayanan2025bias} introduce a real-world news-image benchmark to study social bias in vision--language models, using an LLM-as-a-judge to jointly assess accuracy, faithfulness, and bias. Their findings show that visible social cues can systematically trigger stereotypes, particularly for gender and occupation, and that higher faithfulness does not necessarily correspond to lower bias.

Overall, prior work establishes LLM-as-a-judge as a viable and scalable evaluation paradigm for text-based tasks, while also revealing important limitations related to bias, reliability, and robustness. In contrast, the application of LLM judges to vision and multimodal settings remains comparatively underexplored, with existing studies largely focusing on benchmarking or narrowly scoped safety evaluations. This gap highlights the need for further investigation into how LLM-based judges behave when evaluating structured visual outputs in real-world, safety-critical scenarios.

\section{Methodology}
This section describes the proposed evaluation methodology for assessing the reliability of the LLM-as-Judge in safety-critical inspection scenarios. Rather than focusing on segmentation accuracy, the methodology characterizes the judge’s behavior through two complementary properties: repeatability and sensitivity. Repeatability evaluates the stability of the judge’s numeric outputs under identical inputs and fixed inference settings, while sensitivity examines whether its judgments respond coherently to controlled degradations of visual evidence. Together, these analyses establish whether the LLM-as-Judge can function as a dependable semantic safety monitor in real-world deployment conditions.

\subsection{\textbf{Repeatability and Stability of the LLM-as-Judge}}
\label{sec:repeatability_theory}

Large multimodal language models exhibit strong visual reasoning capabilities, yet their decoding processes are inherently stochastic. In safety-critical applications such as autonomous aerial inspection, this stochasticity introduces unacceptable uncertainty: identical visual inputs must consistently yield identical safety judgments. If an evaluator oscillates between different assessments under identical conditions, it cannot be reliably integrated into downstream decision-making. Consequently, the first requirement of an LLM-as-Judge is Stability, defined here as stable numeric behavior under fixed inputs and inference settings.

We formalize the LLM-as-Judge as a function
\begin{equation}
f_{\theta}(x_i, p) = (s_i, c_i, e_i),
\end{equation}
where $x_i$ denotes an input overlay image composed of an RGB frame and its predicted segmentation mask, and $p$ is a fixed structured prompt. The outputs consist of a discrete segmentation-quality score $s_i \in \{1,\ldots,5\}$, a continuous confidence value $c_i \in [0,1]$, and a textual explanation $e_i$. In deployment, only $(s_i, c_i)$ are used for safety decisions, while $e_i$ provides interpretability and traceability.
\begin{figure*}[t]
\centering
\setlength{\tabcolsep}{2pt}

\newcolumntype{C}[1]{>{\centering\arraybackslash}p{#1}}
\begin{tabular}{C{0.155\textwidth} C{0.155\textwidth} C{0.155\textwidth} C{0.155\textwidth} C{0.155\textwidth} C{0.155\textwidth}}
\textbf{Clean} & \textbf{Rain} & \textbf{Fog} & \textbf{Shadow} & \textbf{Sunflare} & \textbf{Snow} \\[2mm]

\includegraphics[width=\linewidth]{ 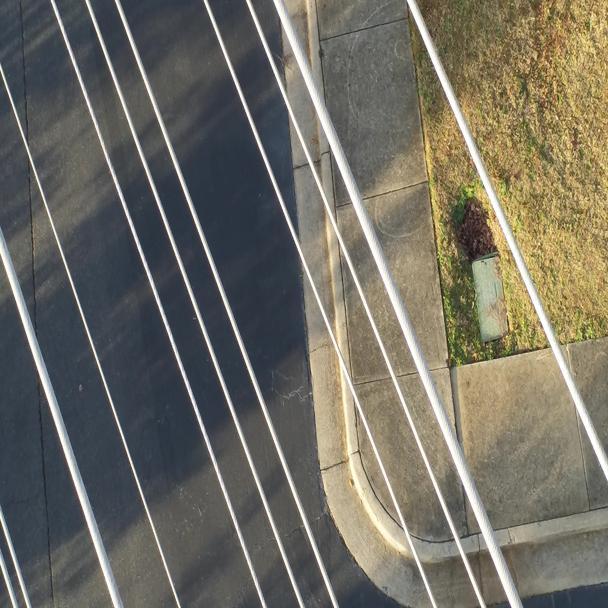} &
\includegraphics[width=\linewidth]{ 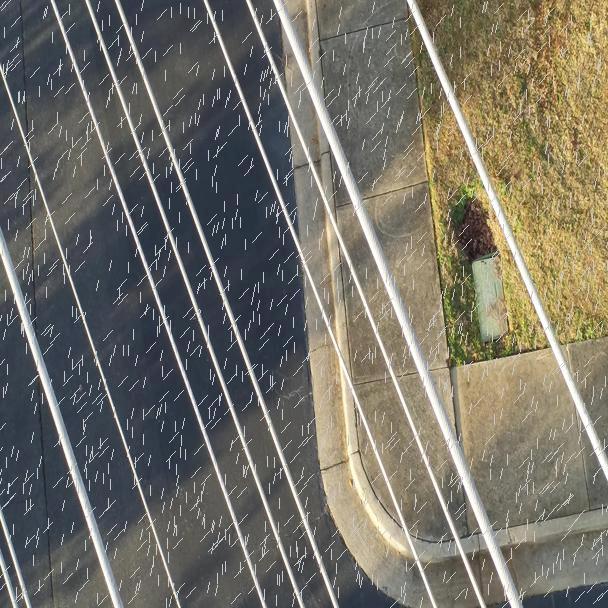} &
\includegraphics[width=\linewidth]{ 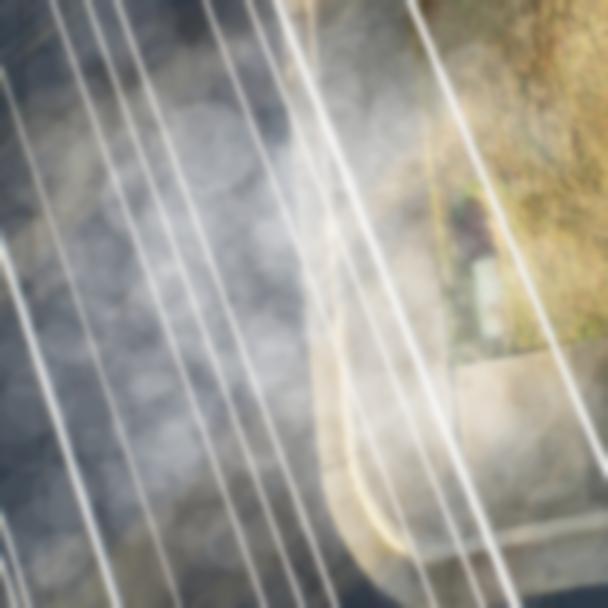} &
\includegraphics[width=\linewidth]{ 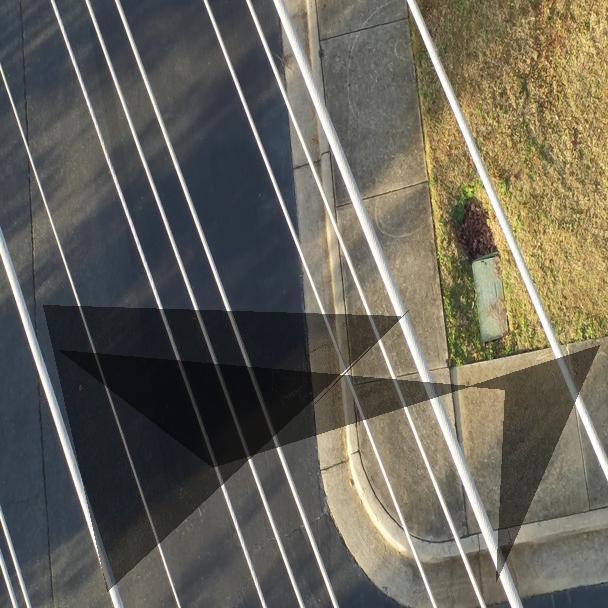} &
\includegraphics[width=\linewidth]{ 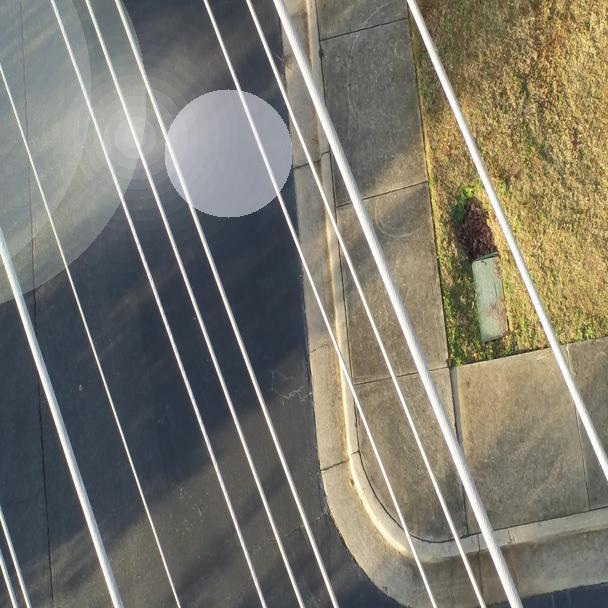} &
\includegraphics[width=\linewidth]{ 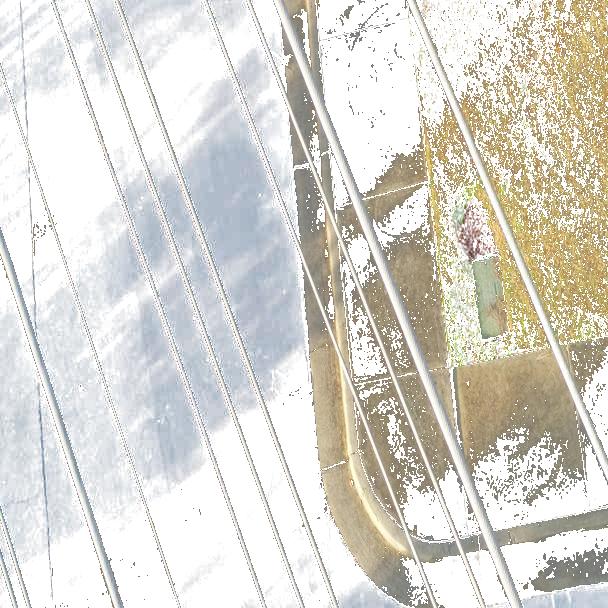} \\[2mm]

\includegraphics[width=\linewidth]{ 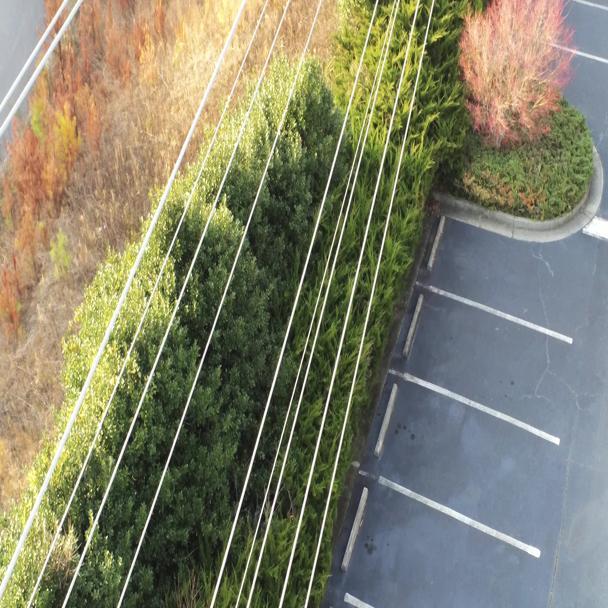} &
\includegraphics[width=\linewidth]{ 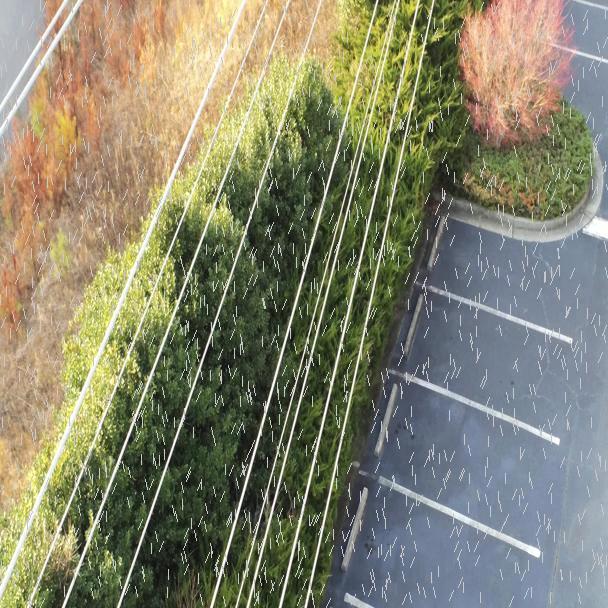} &
\includegraphics[width=\linewidth]{ 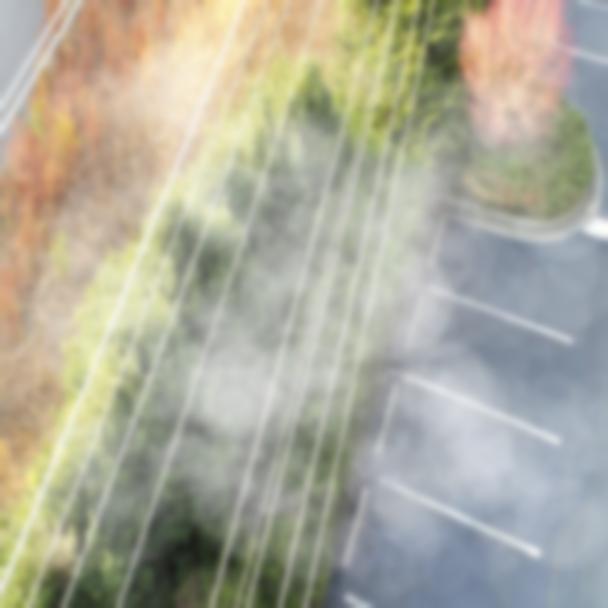} &
\includegraphics[width=\linewidth]{ 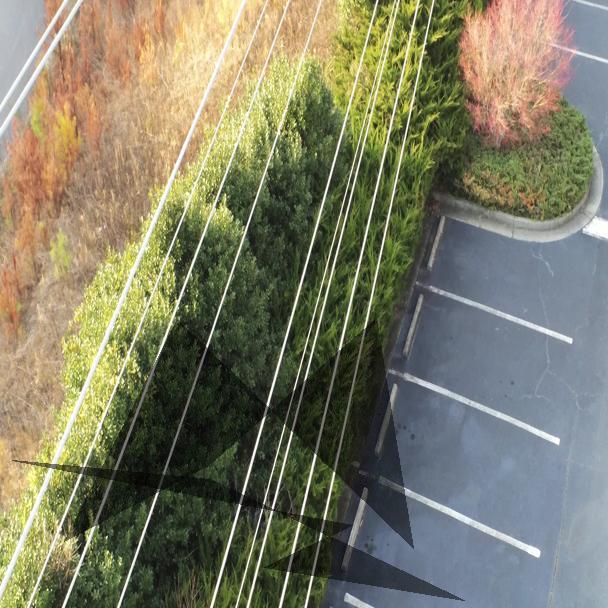} &
\includegraphics[width=\linewidth]{ 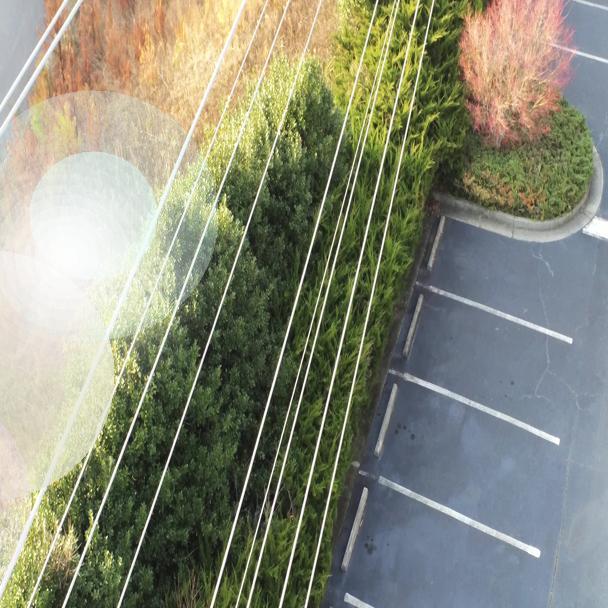} &
\includegraphics[width=\linewidth]{ 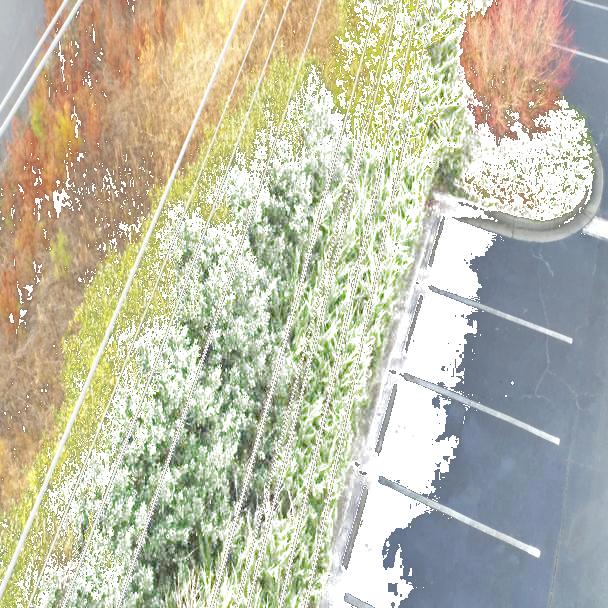} \\

\includegraphics[width=\linewidth]{ 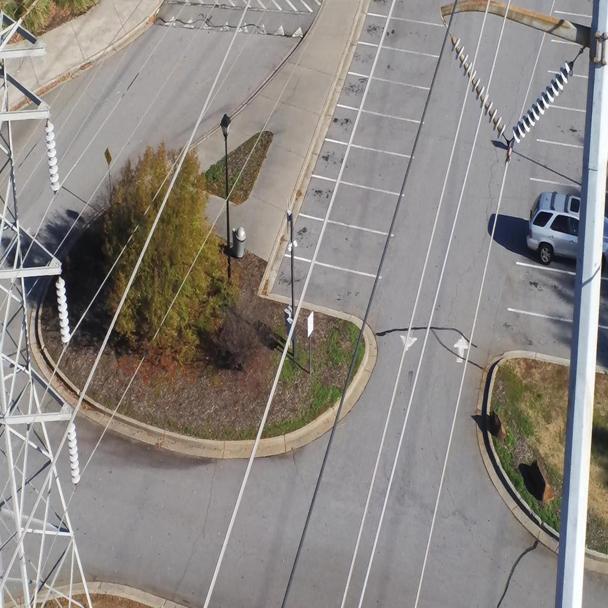} &
\includegraphics[width=\linewidth]{ 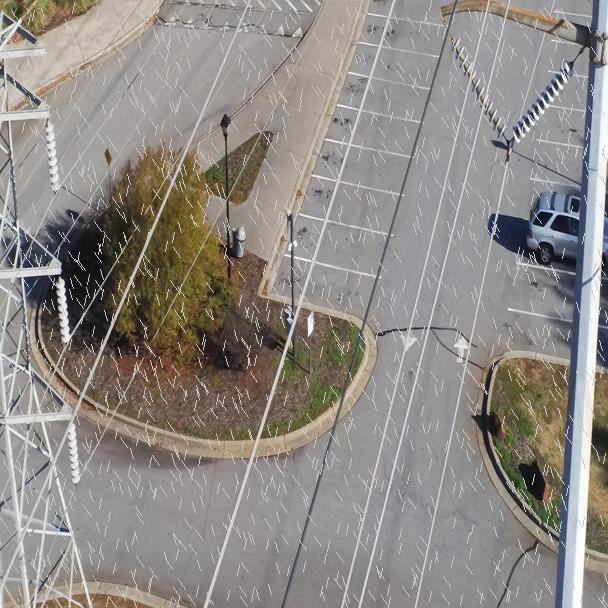} &
\includegraphics[width=\linewidth]{ 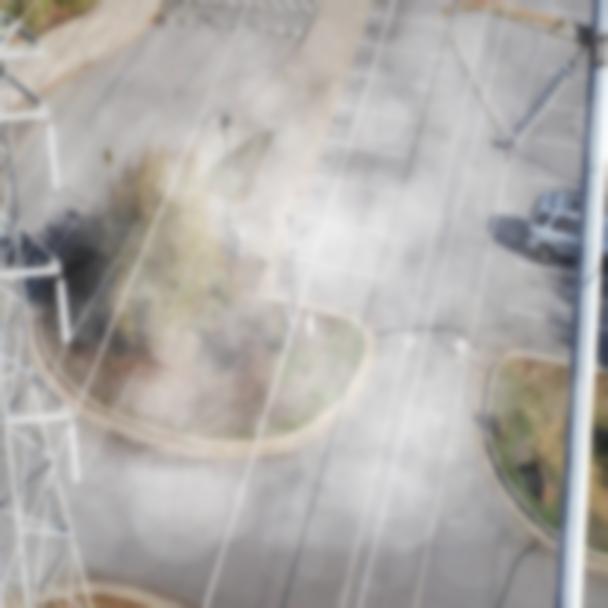} &
\includegraphics[width=\linewidth]{ 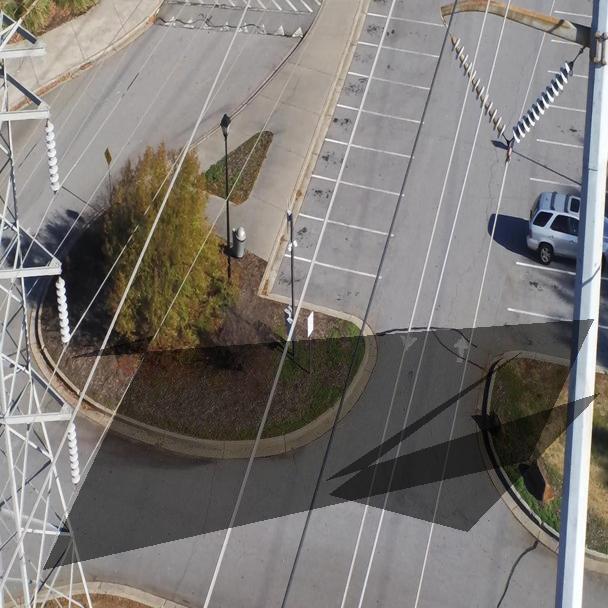} &
\includegraphics[width=\linewidth]{ 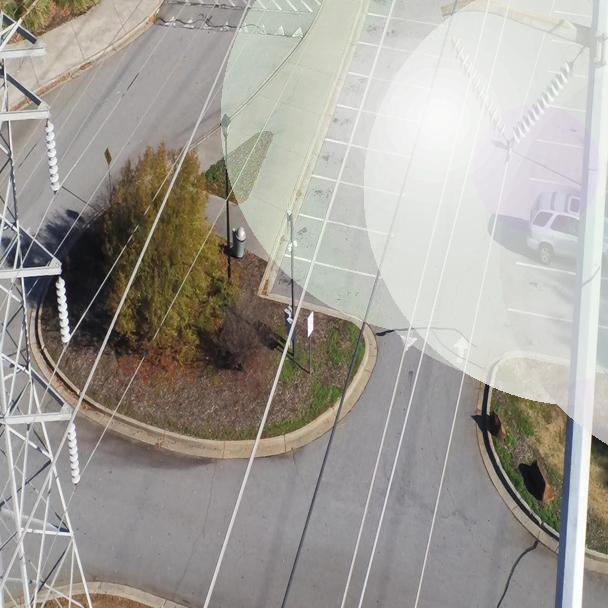} &
\includegraphics[width=\linewidth]{ 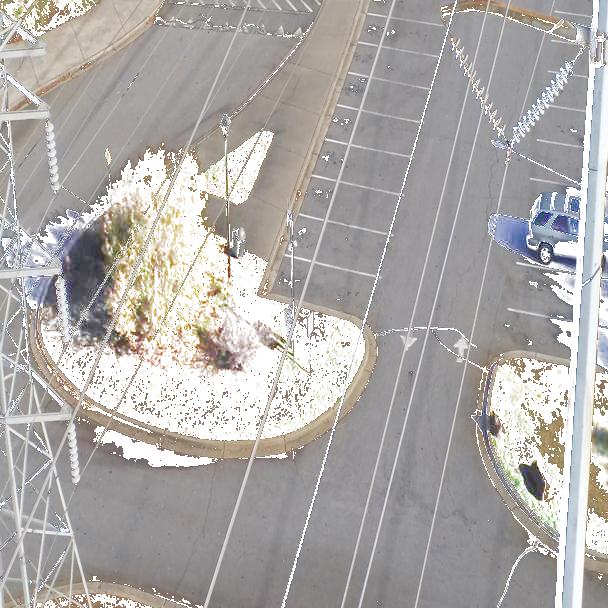} \\
\end{tabular}

\vspace{1mm}
\caption{Examples of clean images and synthetically corrupted variants used to construct the challenge set. Each column corresponds to a corruption type and each row shows a different image instance.}
\label{fig:challenge_examples}
\end{figure*}
Repeatability is evaluated by running the same model multiple times on an identical dataset using the same prompt and decoding parameters. For each image $x_i$, we obtain
\begin{equation}
\{(s_i^{(r)}, c_i^{(r)}, e_i^{(r)})\}_{r=1}^{R},
\end{equation}
where $R$ denotes the number of independent runs. A numerically deterministic judge would satisfy $(s_i^{(r)}, c_i^{(r)}) = (s_i^{(1)}, c_i^{(1)})$ for all $r > 1$, while allowing natural linguistic variation in $e_i^{(r)}$.

The primary indicator of score stability is the score agreement
\begin{equation}
\label{eq:score_agreement}
A_s = \frac{1}{N} \sum_{i=1}^{N} 
\mathbb{I}\!\left(
\max_r s_i^{(r)} - \min_r s_i^{(r)} = 0
\right),
\end{equation}
where $\mathbb{I}(\cdot)$ denotes the indicator function and $N$ denotes the total number of evaluated input samples (overlay images) in the dataset. A high value of $A_s$ indicates that the model assigns the same categorical interpretation of segmentation quality on every invocation and measures the fraction of images receiving identical scores across all runs.

Similarly, confidence agreement is defined as
\begin{equation}
\label{eq:confidence agreement}
A_c = \frac{1}{N} \sum_{i=1}^{N} 
\mathbb{I}\!\left(
\max_r c_i^{(r)} - \min_r c_i^{(r)} \le \epsilon
\right),
\end{equation}
where $\epsilon$ is a small tolerance to account for floating-point variability in probabilistic outputs. In our experiments, we set $\epsilon = 10^{-6}$, enforcing a strict equality criterion that reflects the expectation that confidence outputs remain numerically stable under identical inputs.
A high $A_c$ implies that the model consistently assigns the same degree of confidence to its judgment.
Because downstream safety logic depends on both score and confidence, we define the combined numeric stability
\begin{equation}
\label{eq:combined}
\begin{aligned}
A_{s,c} = \frac{1}{N} \sum_{i=1}^{N} \mathbb{I}\Big(
& \max_{r} s_i^{(r)} - \min_{r} s_i^{(r)} = 0 \\
& \wedge\; \max_{r} c_i^{(r)} - \min_{r} c_i^{(r)} \le \epsilon
\Big).
\end{aligned}
\end{equation}
which represents the probability that both numeric outputs are simultaneously reproducible.

To capture reproducibility beyond strict equality, we additionally report the one-way random-effects intraclass correlation coefficient computed over the score values:
\begin{equation}
\label{eq:icc}
\mathrm{ICC}(1,1) =
\frac{\sigma^2_{\mathrm{between}} - \sigma^2_{\mathrm{within}}}
{\sigma^2_{\mathrm{between}} + (R - 1)\sigma^2_{\mathrm{within}}},
\end{equation}
where $\mathrm{ICC}(1,1)$ denotes the one-way random-effects intraclass correlation coefficient,
$\sigma^2_{\text{between}}$ denotes the variance of the score $s_i$ across different images,
$\sigma^2_{\text{within}}$ denotes the variance of $s_i$ across repeated runs for the same image,
and $R$ is the number of repeated evaluations.

Finally, textual reproducibility is quantified using word-level overlap rather than exact string equality:
\begin{equation}
A_t = \frac{1}{N} \sum_{i=1}^{N} 
\mathrm{Overlap}(e_i^{(1)}, \ldots, e_i^{(R)}),
\end{equation}
where $\mathrm{Overlap}(\cdot)$ denotes the average normalized word overlap across runs for a given image. This metric reflects semantic consistency while accommodating natural linguistic variation. 

\textit{Interpretation of Repeatability Metrics.}
A repeatable judge should exhibit low variability across repeated evaluations of the same image while preserving variability across different images. The intraclass correlation coefficient quantifies this tradeoff by measuring the proportion of total variance attributable to differences between images rather than stochastic fluctuations across runs. High ICC values therefore indicate strong repeatability of the LLM-as-Judge under identical inputs.

\subsection{\textbf{Sensitivity Framework}}
\label{sec:sensitivity}

While repeatability ensures stability under identical conditions, it does not guarantee that the LLM-as-Judge responds meaningfully to changes in visual evidence. A model may be deterministic yet perceptually inert. For reliable deployment, the judge must therefore exhibit sensitivity: its numeric outputs should vary coherently in response to degradations in image quality and structural fidelity.

Sensitivity is evaluated by introducing controlled perturbations that emulate realistic distributional shifts encountered during aerial inspection. For each corruption type and severity level, the degradation is applied to the RGB image prior to segmentation. The corrupted image is then processed by the segmentation network, yielding a correspondingly degraded prediction mask. The corrupted RGB image and its predicted mask are combined into an overlay, which serves as the visual input evaluated by the LLM-as-Judge. All evaluations use the same prompt and inference settings to isolate the effect of visual degradation.

Let $x_{i,0}$ denote the overlay generated from an uncorrupted RGB image and its predicted segmentation mask. For each perturbation type $t$ and severity level $k \in \{1,2,3\}$, the corresponding input $x_{i,t,k}$ is defined as the overlay generated from the corrupted RGB image and its predicted mask. The LLM-as-Judge produces outputs $(s_{i,t,k}^{(r)}, c_{i,t,k}^{(r)}, e_{i,t,k}^{(r)})$ for each input, where the superscript $(r)$ denotes repeated evaluations under identical inference settings. Sensitivity is quantified by comparing these outputs to their clean-reference counterparts.

The mean deviations in score and confidence are defined as
\begin{equation}
\label{eq:mean_deviations}
\begin{aligned}
\Delta s_{t,k} &= \frac{1}{N} \sum_{i=1}^{N} \left( s_{i,0}^{(r)} - s_{i,t,k}^{(r)} \right), \\
\Delta c_{t,k} &= \frac{1}{N} \sum_{i=1}^{N} \left( c_{i,0}^{(r)} - c_{i,t,k}^{(r)} \right).
\end{aligned}
\end{equation}
Unless otherwise stated, the deviations are computed per run and then averaged across repeated evaluations.

To assess whether the response varies monotonically with increasing degradation severity, we compute the Spearman rank correlation between severity level and mean score drop:
\begin{equation}
\rho_{s,t} = \mathrm{Spearman}\left([1,2,3], [\Delta s_{t,1}, \Delta s_{t,2}, \Delta s_{t,3}]\right).
\end{equation}
The Spearman rank correlation measures the strength of a monotonic relationship between two variables based on their ranked values rather than their absolute magnitudes. Unlike Pearson correlation, it does not assume linearity or normally distributed variables, making it well suited for assessing whether the judge’s scores change consistently with increasing degradation severity, regardless of the exact scale or rate of change.
A high value of $\rho_{s,t}$ approaching $1$ indicates that the judge’s scores vary monotonically with increasing perturbation severity, evidencing a coherent perceptual response to progressive visual degradation.

Finally, to determine whether the observed deviations are statistically meaningful at the image level, score residuals are defined as
\begin{equation}
\label{eq:core_residuals}
d_{i,t,k}^{(r)} = s_{i,0}^{(r)} - s_{i,t,k}^{(r)}.
\end{equation}
Paired significance testing is performed on $d_{i,t,k}^{(r)}$ under the null hypothesis
\begin{equation}
\label{eq:null}
H_0 : \mathbb{E}[d_{t,k}] = 0,
\end{equation}
using parametric or non-parametric tests depending on normality, to determine whether the observed deviations arise from systematic perceptual shifts rather than random noise.
\begin{table*}[t]
\centering
\caption{Repeatability metrics over 5 runs under different segmentation corruption conditions (schema-constrained).}
\label{tab:repeatability}
\begin{subtable}[t]{0.33\textwidth}
\centering
\caption{Original}
\vspace{-4pt}
\begin{tabular}{l c}
\hline
Metric & Value \\
\hline
Confidence agreement & 70.05 \\
Score agreement & 81.11 \\
ICC(1,1) & 0.858 \\
Conf. std (mean) & 0.0093 \\
Conf. std (95th) & 0.0424 \\
Text overlap & 40.92 \\
Combined numeric stability ($A_{s,c}$) & 69.59 \\
\hline
\end{tabular}
\end{subtable}
\hfill
\begin{subtable}[t]{0.33\textwidth}
\centering
\caption{Sunflare}
\vspace{-4pt}
\begin{tabular}{l c}
\hline
Metric & Value \\
\hline
Confidence agreement & 60.83 \\
Score agreement & 82.49 \\
ICC(1,1) & 0.901 \\
Conf. std (mean) & 0.0151 \\
Conf. std (95th) & 0.0548 \\
Text overlap & 39.58 \\
Combined numeric stability ($A_{s,c}$)  & 60.37 \\
\hline
\end{tabular}
\end{subtable}
\begin{subtable}[t]{0.33\textwidth}
\centering
\caption{Rain}
\vspace{-4pt}
\begin{tabular}{l c}
\hline
Metric & Value \\
\hline
Confidence identical (all 5 runs) & 58.53 \\
Score agreement (all 5 runs) & 82.95 \\
ICC(1,1) for scores & 0.888 \\
Confidence std (mean) & 0.0132 \\
Confidence std (95th pct.) & 0.0548 \\
Text explanation word-overlap & 44.53 \\
Combined numeric stability ($A_{s,c}$) & 58.53 \\
\hline
\end{tabular}
\end{subtable}
\hfill
\begin{subtable}[t]{0.33\textwidth}
\centering
\vspace{4pt}
\caption{Shadow}
\vspace{-4pt}
\begin{tabular}{l c}
\hline
Metric & Value \\
\hline
Confidence agreement & 54.38 \\
Score agreement & 80.18 \\
ICC(1,1) & 0.880 \\
Conf. std (mean) & 0.0179 \\
Conf. std (95th) & 0.0572 \\
Text overlap & 37.74 \\
Combined numeric stability ($A_{s,c}$) & 53.46 \\
\hline
\end{tabular}
\end{subtable}
\hfill
\begin{subtable}[t]{0.33\textwidth}
\centering
\vspace{4pt}
\caption{Snow}
\vspace{-4pt}
\begin{tabular}{l c}
\hline
Metric & Value \\
\hline
Confidence agreement & 43.78 \\
Score agreement & 78.80 \\
ICC(1,1) & 0.898 \\
Conf. std (mean) & 0.0225 \\
Conf. std (95th) & 0.0707 \\
Text overlap & 36.72 \\
Combined numeric stability ($A_{s,c}$) & 42.86 \\
\hline
\end{tabular}
\end{subtable}
\hfill
\begin{subtable}[t]{0.33\textwidth}
\centering
\vspace{4pt}
\caption{Fog}
\vspace{-4pt}
\begin{tabular}{l c}
\hline
Metric & Value \\
\hline
Confidence agreement & 33.18 \\
Score agreement & 90.78 \\
ICC(1,1) & 0.917 \\
Conf. std (mean) & 0.0880 \\
Conf. std (95th) & 0.3162 \\
Text overlap & 68.11 \\
Combined numeric stability ($A_{s,c}$) & 33.18 \\
\hline
\end{tabular}
\end{subtable}
\end{table*}

\section{Experiment Setup}
\subsection{Dataset preparation}
We conduct our experiments using the TTPLA dataset~\cite{abdelfattah2020ttpla}, a public benchmark for powerline inspection that contains aerial RGB images captured from diverse viewpoints, altitudes, and backgrounds. The dataset includes approximately 1{,}100 images with fine-grained polygon annotations of powerline structures, where powerlines occupy a small fraction of the image area, reflecting the inherent sparsity and thin geometry of the target objects. 

To evaluate the robustness of the proposed framework under adverse operating conditions, we construct a \emph{challenge set} derived from the original TTPLA images. Starting from the clean RGB images, we apply synthetic corruptions using the \texttt{Albumentations} library to simulate visual degradations commonly encountered during aerial inspection. These corruptions introduce controlled variations in visibility and illumination while preserving pixel-level alignment between clean and perturbed images.

Each corruption type is applied at three severity levels ($k \in \{1,2,3\}$). The considered corruption families include (i) \textbf{weather effects}, such as fog, rain, snow, and sun flare, which reduce global contrast and obscure line continuity, and (ii) \textbf{lighting shifts}, implemented through shadow perturbations that locally alter luminance and visibility of thin structures. This process yields a balanced and visually diverse challenge set without introducing artificial artifacts that could confound the semantic reasoning of the LLM.

\subsection{LLM-as-Judge Configuration.}
All experiments use a single multimodal large language model as the LLM-as-Judge. Specifically, we employ \textit{GPT-4o}~\cite{openai_gpt4o}, accessed via the OpenAI API, which supports image--text inputs and exhibits strong visual reasoning capabilities required for evaluating segmentation overlays. The model receives RGB--mask overlay images together with a fixed structured prompt and produces a discrete quality score, a confidence estimate, and a textual rationale.

\subsection{Experiment Setup for Repeatability}

To evaluate repeatability, we conduct experiments on both the original TTPLA dataset and the constructed challenge set. Each RGB image, whether clean or corrupted, is processed by a U-Net model trained for 25 epochs on the original TTPLA training split to generate a predicted segmentation mask. The resulting mask–image overlays are then evaluated by the LLM-as-Judge, which assigns a discrete quality score and confidence estimate to each sample. This evaluation is repeated over five independent runs using identical inputs and fixed inference settings, allowing us to assess the stability of the judge’s outputs across clean images and all corruption conditions.

\subsection{Experiment Setup for Sensitivity}

For sensitivity analysis, we evaluate the LLM-as-Judge on the clean TTPLA images and on all corrupted variants across three severity levels. For each condition, the RGB image is first segmented by the U-Net model, and the resulting mask–image overlay is provided to the LLM-as-Judge. For every sample, the judge produces a quality score, a confidence estimate, a textual explanation, and a response latency (ms), enabling analysis of how the judge’s outputs vary in response to increasing visual degradation.
\begin{table*}[!t]
\centering
\caption{Sensitivity Statistics of LLM-as-Judge Scores and Confidence Under Controlled Visual Corruptions}
\resizebox{\textwidth}{!}{%
\begin{tabular}{lrrrrrrrrrrrr}
\toprule
corruption & severity & mean\_ds & std\_ds & ci95\_ds\_lo & ci95\_ds\_hi & mean\_dc & std\_dc & ci95\_dc\_lo & ci95\_dc\_hi & dz\_score & dz\_conf \\
\midrule
fog & 1 & 3.124 & 0.912 & 2.991 & 3.249 & 0.736 & 0.213 & 0.705 & 0.762 & 3.426 & 3.453 \\
fog & 2 & 3.143 & 0.949 & 3.005 & 3.263 & 0.741 & 0.211 & 0.711 & 0.768 & 3.311 & 3.509 \\
fog & 3 & 3.115 & 0.918 & 2.982 & 3.235 & 0.732 & 0.214 & 0.701 & 0.759 & 3.393 & 3.419 \\
rain & 1 & 0.465 & 0.828 & 0.350 & 0.571 & 0.052 & 0.149 & 0.030 & 0.071 & 0.562 & 0.346 \\
rain & 2 & 0.581 & 0.760 & 0.479 & 0.677 & 0.067 & 0.120 & 0.050 & 0.082 & 0.764 & 0.558 \\
rain & 3 & 0.806 & 0.871 & 0.687 & 0.922 & 0.109 & 0.166 & 0.085 & 0.131 & 0.926 & 0.654 \\
shadow & 1 & 0.631 & 0.741 & 0.530 & 0.733 & 0.071 & 0.117 & 0.054 & 0.086 & 0.852 & 0.604 \\
shadow & 2 & 0.581 & 0.863 & 0.465 & 0.700 & 0.063 & 0.147 & 0.041 & 0.082 & 0.673 & 0.428 \\
shadow & 3 & 0.562 & 0.820 & 0.452 & 0.673 & 0.056 & 0.146 & 0.035 & 0.076 & 0.685 & 0.386 \\
snow & 1 & 0.700 & 0.975 & 0.567 & 0.834 & 0.101 & 0.207 & 0.074 & 0.130 & 0.718 & 0.489 \\
snow & 2 & 0.853 & 1.026 & 0.710 & 0.991 & 0.130 & 0.222 & 0.101 & 0.160 & 0.831 &  0.586 \\
snow & 3 & 0.963 & 1.004 & 0.825 & 1.101 & 0.147 & 0.227 & 0.118 & 0.177 & 0.959 & 0.647 \\
sunflare & 1 & 0.479 & 0.811 & 0.373 & 0.590 & 0.054 & 0.145 & 0.034 & 0.073 & 0.591 & 0.371 \\
sunflare & 2 & 0.558 & 0.798 & 0.452 & 0.664 & 0.067 & 0.146 & 0.046 & 0.085 & 0.699 & 0.457 \\
sunflare & 3 & 0.507 & 0.788 & 0.401 & 0.608 & 0.056 & 0.131 & 0.037 & 0.072 & 0.643 & 0.425 \\
\bottomrule
\end{tabular}%
}
\label{tab:sensitivity_result}
\end{table*}
\section{Results Analysis}
\subsection{Repeatability and Stability of LLM-as-Judge}
Table~\ref{tab:repeatability} extend our repeatability study from the clean TTPLA\cite{abdelfattah2020ttpla} images to the corrupted challenge set, showing how different real-world conditions affect the stability of the LLM-as-Judge.  
The main divergences appear in the confidence agreement $A_c$ (Eq.~(\ref{eq:confidence agreement})). Depending on the corruption, the confidence-identical rate varies widely from relatively stable behavior under sunflare (60.83\%) and rain (58.53\%) to much lower stability under fog (33.18\%) and snow (43.78\%). These patterns match the confidence-standard-deviation results: fog, in particular, introduces strong variation in confidence, which is expected because the blur and haze obscure powerlines and remove reliable visual cues. Conversely, corruptions like rain or shadow degrade the scene without fully eliminating structure, allowing the LLM to maintain more stable uncertainty estimates.

The combined numeric stability follows a similar trend. It is high on clean images (69.59\%) but drops sharply under fog (33.18\%), showing that while discrete scoring remains stable, confidence calibration becomes more sensitive to loss of visual evidence. This joint behavior is captured by the combined numeric stability metric $A_{s,c}$ defined in Eq.~(\ref{eq:combined}), which enforces simultaneous consistency of both score and confidence outputs under identical inputs.
From a safety perspective, this is desirable: when cues degrade, the judge appropriately becomes less certain, behaving as a cautious evaluator rather than over-confidently hallucinating structure. As expected, textual-explanation match rates stay low across all settings, since natural linguistic variation does not affect the underlying semantic judgment.
    
    

Overall, the model remains quite consistent in its discrete scoring: across all corruption types, the score agreement $A_s$ defined in Eq.~(\ref{eq:score_agreement}) remains between 78--91\%. At first glance, it may seem counterintuitive that some corrupted settings, such as fog (90.78\%) and snow (78.80\%), exhibit even higher score agreement than the clean images (81.11\%). However, this trend aligns with how corruption interacts with the underlying segmentation model. In many fog- and snow-corrupted cases, the segmentation model fails to detect most or all powerlines, producing overlays with few or no highlighted structures. These ``empty'' or near-empty overlays lead to simple, uniformly poor segmentation cases, which the LLM judge evaluates consistently across runs, typically assigning a stable low score.
In contrast, clean images contain multiple visible powerlines, often appearing in varied positions and levels of local visibility. This forces the LLM to inspect several line instances per image, some of which may be partially occluded or faint. As a result, the judgment task is more complex: subtle variations in attention or phrasing during each run can cause small fluctuations in the assigned score. Thus, clean images feature more semantic load per judgment, which naturally lowers perfect score agreement relative to heavily corrupted cases where the segmentation model often fails more uniformly and predictably.

The ICC values (0.858--0.917), computed using Eq.~(\ref{eq:icc}), reinforce this interpretation by showing that the relative ranking of image quality remains stable across corruption conditions, even when exact score agreement varies. Since ICC captures consistency in ordinal structure rather than exact label matching, these results indicate that the LLM-as-Judge preserves a reliable notion of structural correctness even under changing corruption regimes, which is essential for threshold-based safety monitoring.

Taken together, these findings support the two key ideas introduced in Section~III-A. First, the LLM-as-Judge exhibits strong stability in its discrete quality scores across repeated evaluations, even under significant visual corruption. Second, its confidence estimates adapt proportionally to the degradation of visual evidence, particularly in challenging conditions such as fog and snow where structural cues are severely diminished. This combination of stable categorical judgment and adaptive uncertainty indicates that the LLM-as-Judge functions as a reliable and cautious semantic monitor for real-time powerline inspection. Under fixed inputs and inference settings, the observed numeric stability satisfies a necessary requirement for deployment in safety-critical monitoring pipelines.

\subsection{Sensitivity to Segmentation of the LLM-as-Judge}
Table~\ref{tab:sensitivity_result} summarizes the sensitivity of the LLM-as-Judge to controlled degradations of segmentation overlays across five corruption families and three severity levels. Sensitivity is quantified relative to a clean-reference baseline using the score and confidence deviations defined in Eq.~(\ref{eq:mean_deviations}).
Columns \texttt{mean\_ds} and \texttt{mean\_dc} report the mean score and confidence deviations, $\Delta s_{t,k}$ and $\Delta c_{t,k}$, respectively, measuring the average drop in discrete quality score and confidence under corruption type $t$ and severity level $k$. These values reflect the magnitude of the judge’s perceptual response relative to the uncorrupted condition.
To characterize variability across images, the columns \texttt{std\_ds} and \texttt{std\_dc} report the standard deviations of the per-image residuals defined in Eq.~(\ref{eq:core_residuals}). The corresponding columns \texttt{ci95\_ds\_lo}, \texttt{ci95\_ds\_hi}, \texttt{ci95\_dc\_lo}, and \texttt{ci95\_dc\_hi} provide 95\% confidence intervals computed from these residuals, capturing the consistency of the judge’s response across the dataset.

Finally, the columns \texttt{dz\_score} and \texttt{dz\_conf} report standardized paired-effect sizes computed from the same per-image residuals. These effect sizes correspond to the paired significance testing performed under the null hypothesis in Eq.~(\ref{eq:null}) and quantify whether the observed deviations reflect systematic perceptual sensitivity rather than random stochastic variation.
Together, these statistics provide a complete characterization of sensitivity, capturing not only the magnitude of score and confidence changes under visual degradation, but also their consistency and statistical significance.

Across corruption types, the judge exhibits a coherent sensitivity profile that aligns with the watchdog objective of detecting reliability loss in safety-critical powerline inspection. Fog induces by far the largest and most stable degradation: $\text{mean\_ds} \approx 3.12$--$3.14$ across severities, paired with a large confidence drop of $\text{mean\_dc} \approx 0.73$--$0.74$. The corresponding effect sizes are extremely high for both score and confidence ($d_z \approx 3.31$--$3.43$ and $d_z \approx 3.42$--$3.51$, respectively), indicating strong per-image agreement that fogged overlays represent severe structural unreliability. Notably, the response is nearly saturated across severity levels, suggesting a ceiling effect: once haze substantially obscures thin line evidence and/or causes the segmentation to collapse toward missing structures, increasing fog strength produces only marginal additional perceived loss.

Rain and snow exhibit the expected monotonic sensitivity pattern. For rain, $\text{mean\_ds}$ increases from $0.465$ to $0.806$ as severity rises, with confidence drops increasing from $0.052$ to $0.109$; the paired effect sizes similarly strengthen ($d_z$ for score from $0.562$ to $0.926$, and for confidence from $0.346$ to $0.654$). Snow follows the same trend, with $\text{mean\_ds}$ rising from $0.700$ to $0.963$ and $\text{mean\_dc}$ from $0.101$ to $0.147$, accompanied by increasing effect sizes (score $d_z$ from $0.718$ to $0.959$, confidence $d_z$ from $0.489$ to $0.647$). These results indicate that the judge not only detects worsening segmentation reliability under progressive weather clutter, but does so consistently at the per-image level.

Lighting-driven corruptions (shadow and sunflare) produce smaller and less strictly monotonic shifts, consistent with the fact that such distortions can be locally disruptive without uniformly breaking global line continuity. Shadow shows modest score drops ($\text{mean\_ds}$ decreasing from $0.631$ to $0.562$ across severities) and conservative confidence drops ($\text{mean\_dc}$ from $0.071$ to $0.056$), with positive but moderate effect sizes (score $d_z \approx 0.67$--$0.85$, confidence $d_z \approx 0.39$--$0.60$). Sunflare behaves similarly: while severity 2 yields the largest average drop ($\text{mean\_ds}=0.558$, $\text{mean\_dc}=0.067$), severity 3 partially rebounds ($\text{mean\_ds}=0.507$, $\text{mean\_dc}=0.056$), and effect sizes remain moderate (score $d_z \approx 0.59$--$0.70$, confidence $d_z \approx 0.37$--$0.46$). This mild non-monotonicity is plausible in glare regimes where additional saturation can reduce visible clutter in ways that do not linearly translate to perceived segmentation error, especially when the judge relies on the overlay’s geometric coherence rather than raw photometric fidelity.
\begin{figure}[t]
\centering

\begin{subfigure}{0.43\textwidth}
\includegraphics[width=\linewidth]{ 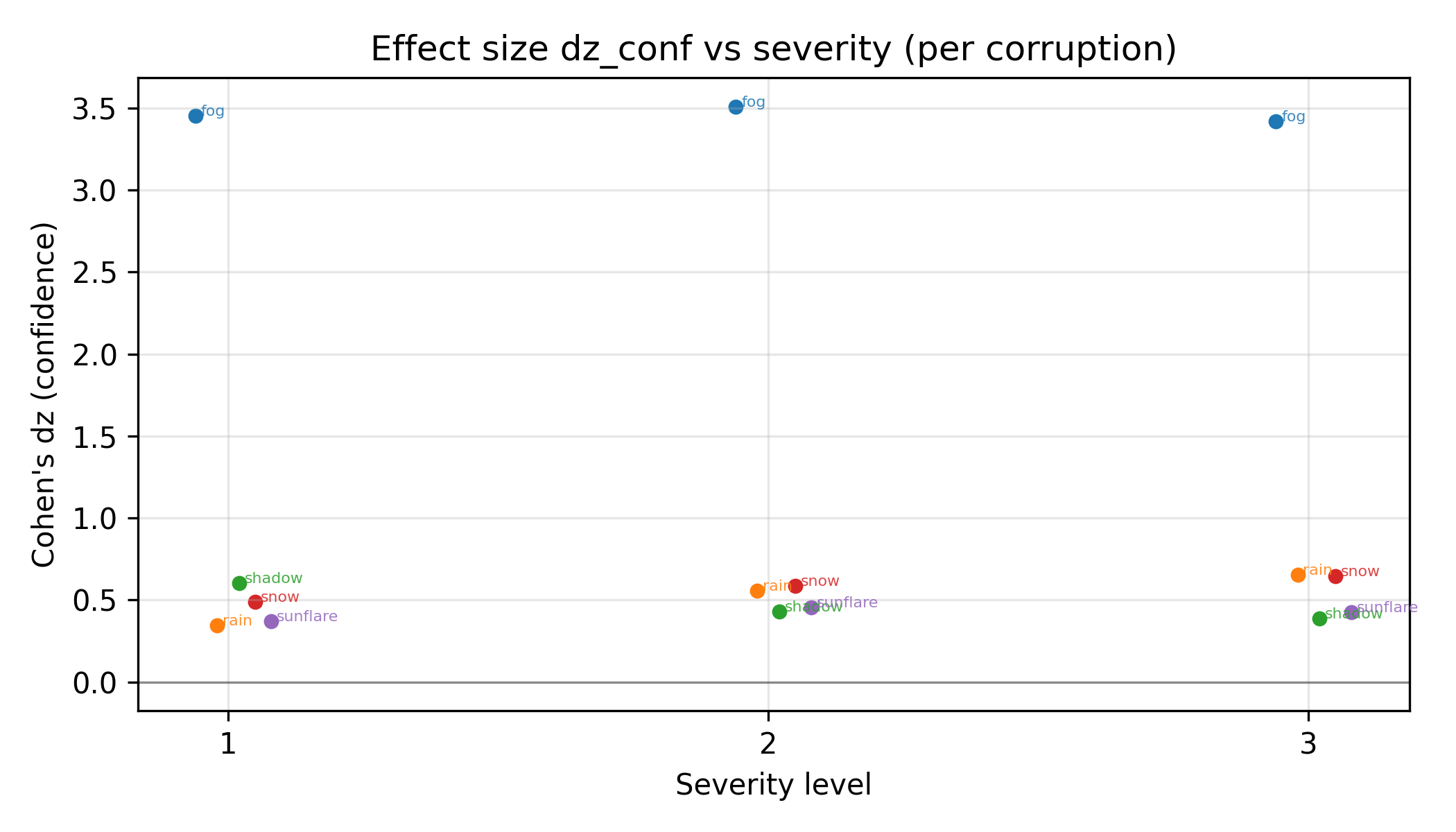}
\caption{Effect size $d_z$ (confidence) vs. severity}
\end{subfigure}
\hfill
\begin{subfigure}{0.43\textwidth}
\includegraphics[width=\linewidth]{ 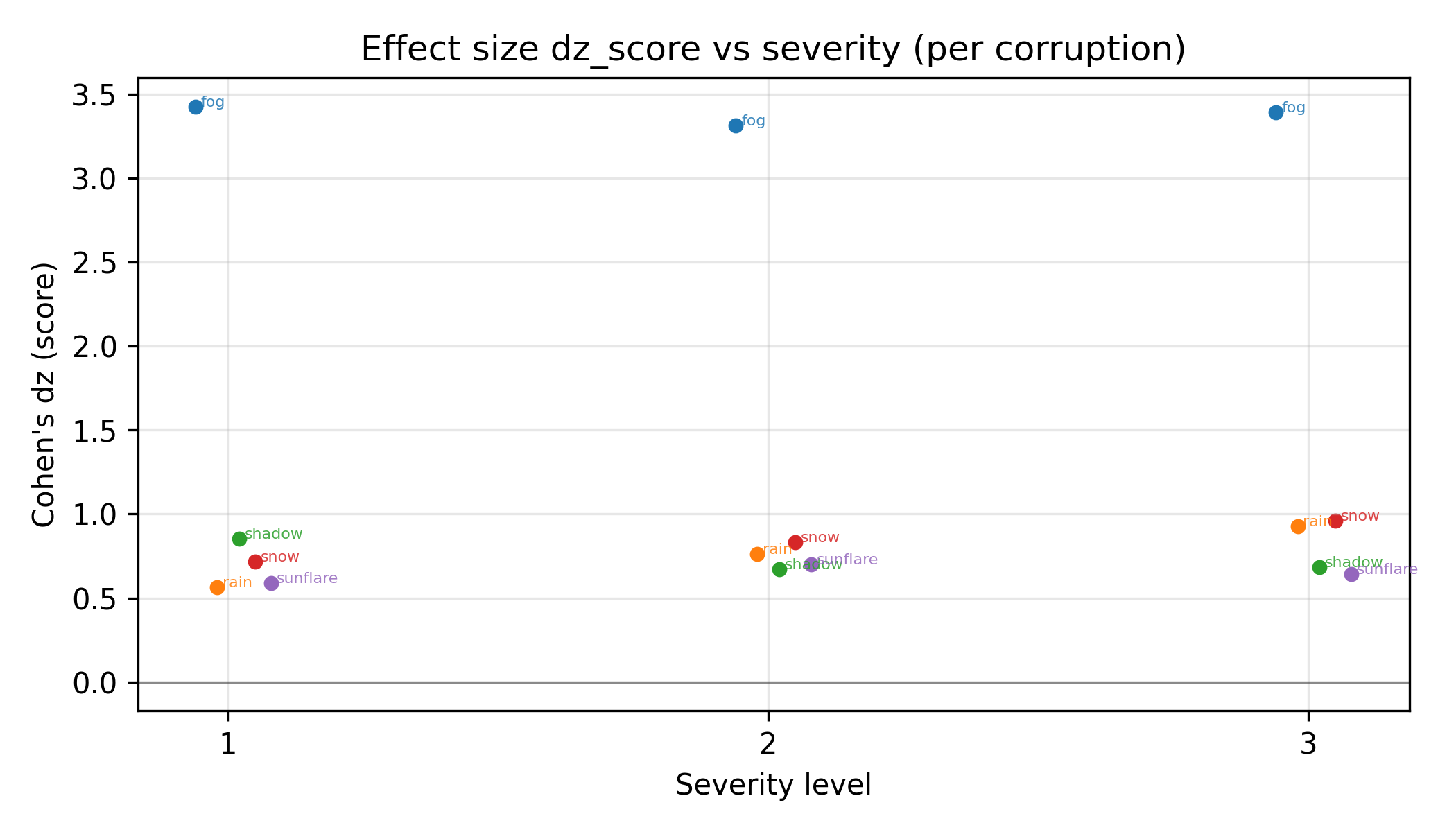}
\caption{Effect size $d_z$ (score) vs. severity}
\end{subfigure}

\vspace{0.3cm}

\begin{subfigure}{0.43\textwidth}
\includegraphics[width=\linewidth]{ 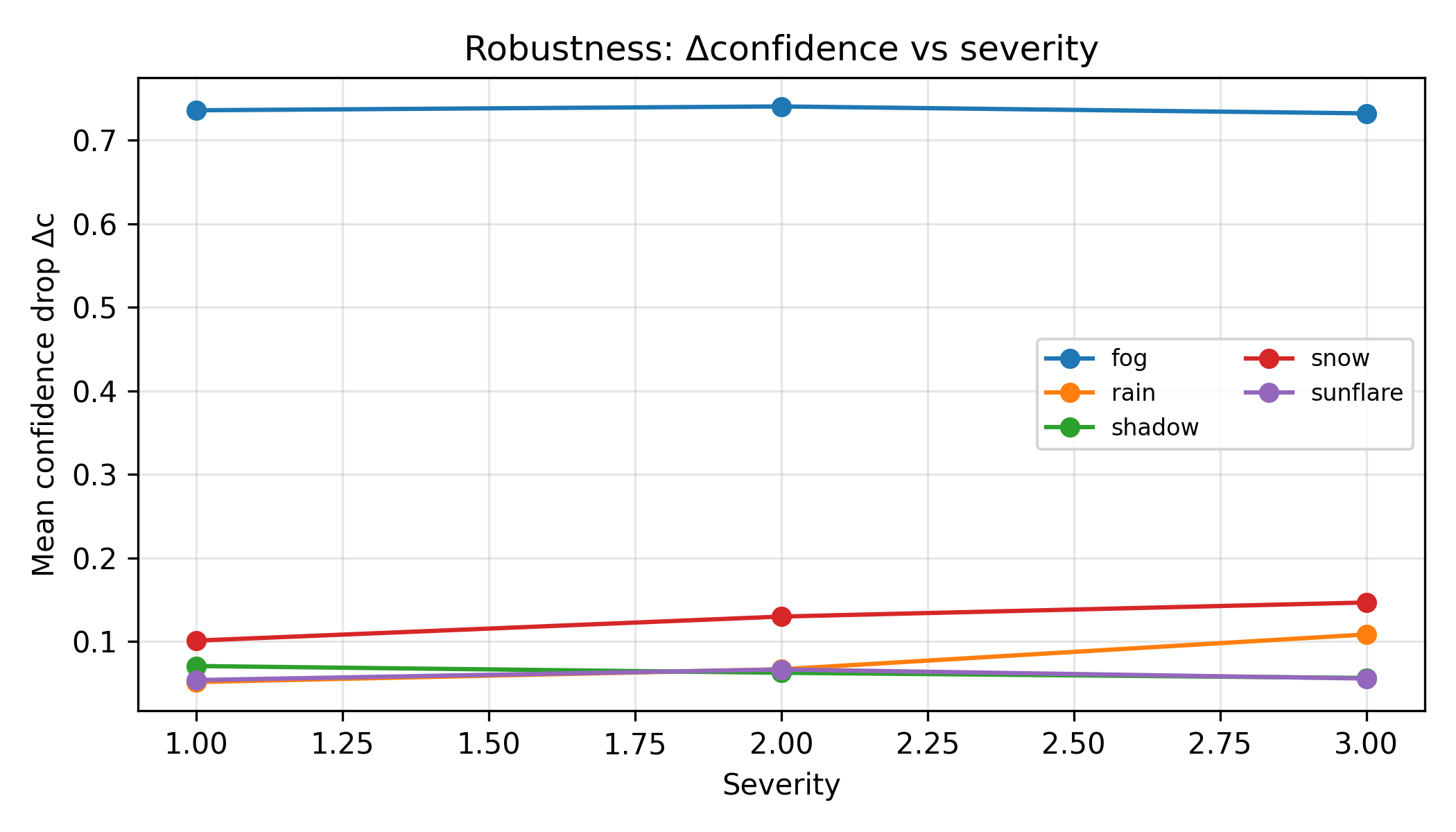}
\caption{Mean confidence drop $\Delta c$ vs. severity}
\end{subfigure}
\hfill
\begin{subfigure}{0.43\textwidth}
\includegraphics[width=\linewidth]{ 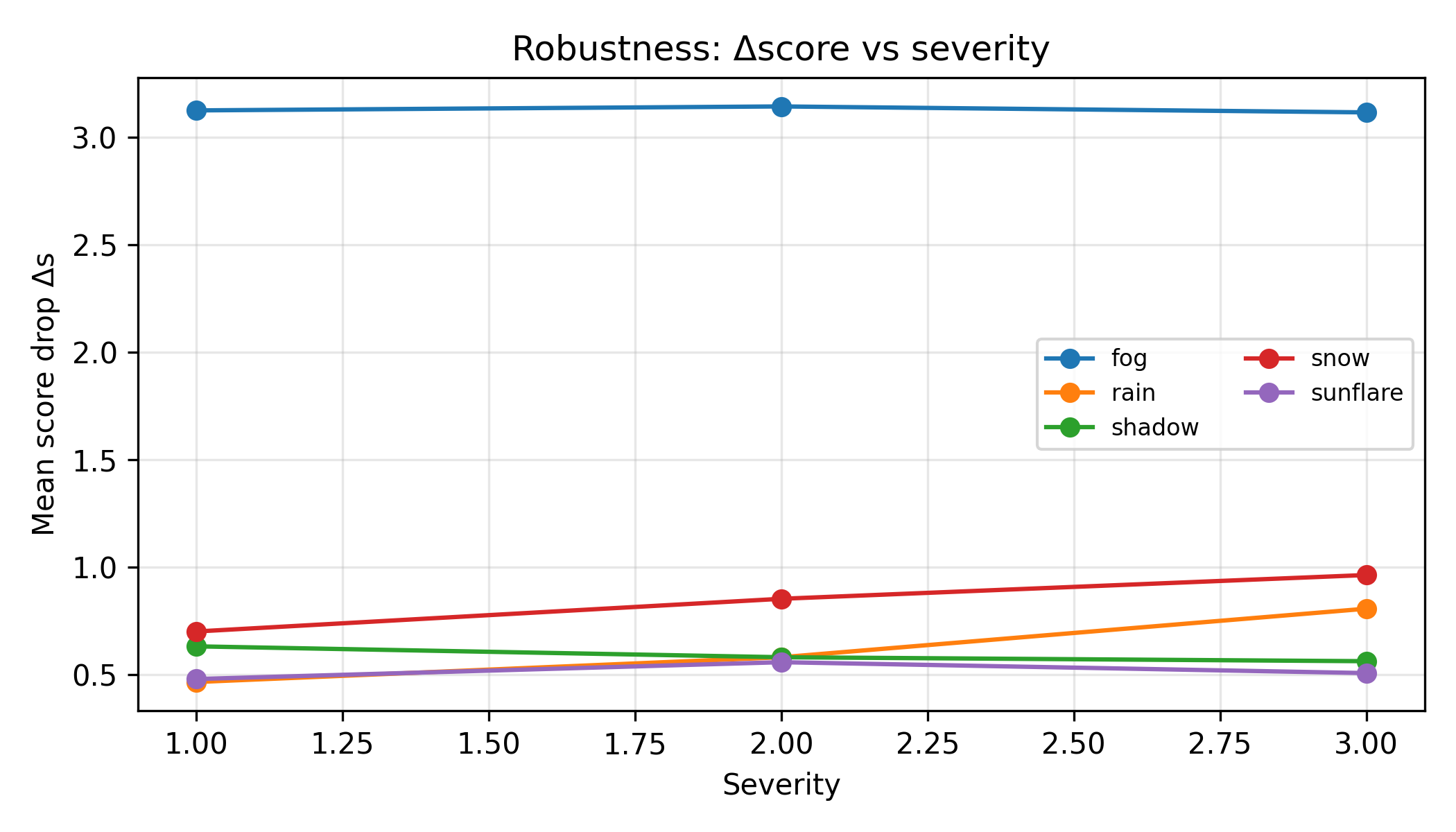}
\caption{Mean score drop $\Delta s$ vs. severity}
\end{subfigure}

\caption{Visualization of sensitivity across corruption types.}
\label{fig:sensitivity_plots}
\end{figure}
Overall, the judge’s confidence signal is systematically more conservative than its discrete score signal for all non-fog corruptions (e.g., $\text{mean\_dc}\le 0.147$ while $\text{mean\_ds}$ approaches $\sim 1.0$ under severe snow), whereas fog triggers large coupled declines in both score and confidence. This is desirable for an automated watchdog: the judge remains discriminative in its quality scoring while reserving strong uncertainty escalation for the most visibility-destroying conditions. Taken together, the monotonic trends for rain and snow, the strong saturation-consistent response for fog, and the consistently positive per-image effect sizes across all conditions support the conclusion that the LLM-as-Judge is perceptually grounded in segmentation reliability cues (e.g., missing or misidentified powerlines) and responds robustly to real-world degradations that threaten safe autonomous inspection.


Figure~\ref{fig:sensitivity_plots} illustrates the sensitivity of the LLM as Judge across corruption types and severity levels. The top row reports per image effect sizes $d_z$ for confidence and score, showing that fog induces the strongest and most consistent perceptual degradation, with very large effect sizes exceeding $3$. Other corruptions, including rain, snow, shadow, and sunflare, exhibit moderate but stable effect sizes, indicating that the judge reliably detects meaningful structural distortions even when degradation is less severe. The bottom row shows the mean drops in score and confidence relative to clean images. Score deviations increase monotonically with severity for rain, snow, and sunflare, reflecting progressive loss of structural fidelity in the predicted masks. Confidence follows the same trend but with smaller magnitudes, demonstrating a conservative and proportional uncertainty response. Fog again produces the largest and most uniform decline across severities, confirming its strong impact on geometric visibility. Together, these results demonstrate that the LLM-as-Judge does not merely produce stable outputs, but responds systematically to meaningful changes in visual evidence, satisfying the sensitivity criterion required for semantic safety monitoring.

\section{Conclusion}

This paper studied the reliability of large multimodal language models when used as automated judges for segmentation quality in safety critical visual inspection tasks. We proposed an evaluation framework that characterizes reliability through two complementary properties, repeatability and sensitivity. Repeatability captures the stability of the judge under identical inputs, while sensitivity evaluates whether the judge responds coherently to controlled degradations in visual evidence. Together, these properties define the behavioral requirements of a trustworthy LLM based evaluator.

Experiments on clean and synthetically corrupted powerline inspection imagery demonstrate that the LLM as Judge exhibits strong repeatability in its discrete quality scores, with high agreement rates and consistent intraclass correlation values across conditions. Confidence estimates decrease appropriately as visual cues deteriorate, particularly under severe corruptions such as fog and snow, indicating adaptive uncertainty rather than overconfident behavior. Sensitivity analysis further shows that score and confidence deviations increase monotonically with corruption severity, supported by large per image effect sizes that reflect meaningful perceptual shifts rather than noise. Together, these results indicate that large multimodal language models can serve as reliable semantic monitors for deployed vision systems when their behavior is rigorously characterized beyond traditional accuracy metrics. 

\bibliographystyle{IEEEtran}
\bibliography{references} 

@inproceedings{li2025generation,
  title={From generation to judgment: Opportunities and challenges of LLM-as-a-judge},
  author={Li, Dawei and Jiang, Bohan and Huang, Liangjie and Beigi, Alimohammad and Zhao, Chengshuai and Tan, Zhen and Bhattacharjee, Amrita and Jiang, Yuxuan and Chen, Canyu and Wu, Tianhao and others},
  booktitle={Proceedings of the 2025 Conference on Empirical Methods in Natural Language Processing},
  pages={2757--2791},
  year={2025}
}

@article{zhu2023judgelm,
  title={{JudgeLM}: Fine-tuned large language models are scalable judges},
  author={Zhu, Lianghui and Wang, Xinggang and Wang, Xinlong},
  journal={arXiv preprint arXiv:2310.17631},
  year={2023}
}

@article{ye2024justice,
  title={Justice or prejudice? quantifying biases in {LLM}-as-a-judge},
  author={Ye, Jiayi and Wang, Yanbo and Huang, Yue and Chen, Dongping and Zhang, Qihui and Moniz, Nuno and Gao, Tian and Geyer, Werner and Huang, Chao and Chen, Pin-Yu and others},
  journal={arXiv preprint arXiv:2410.02736},
  year={2024}
}

@article{zheng2023judging,
  title={Judging {LLM}-as-a-judge with mt-bench and chatbot arena},
  author={Zheng, Lianmin and Chiang, Wei-Lin and Sheng, Ying and Zhuang, Siyuan and Wu, Zhanghao and Zhuang, Yonghao and Lin, Zi and Li, Zhuohan and Li, Dacheng and Xing, Eric and others},
  journal={Advances in neural information processing systems},
  volume={36},
  pages={46595--46623},
  year={2023}
}

@article{pan2024human,
  title={Human-Centered Design Recommendations for LLM-as-a-judge},
  author={Pan, Qian and Ashktorab, Zahra and Desmond, Michael and Cooper, Martin Santillan and Johnson, James and Nair, Rahul and Daly, Elizabeth and Geyer, Werner},
  journal={arXiv preprint arXiv:2407.03479},
  year={2024}
}

@inproceedings{szymanski2025limitations,
  title={Limitations of the LLM-as-a-judge approach for evaluating LLM outputs in expert knowledge tasks},
  author={Szymanski, Annalisa and Ziems, Noah and Eicher-Miller, Heather A and Li, Toby Jia-Jun and Jiang, Meng and Metoyer, Ronald A},
  booktitle={Proceedings of the 30th International Conference on Intelligent User Interfaces},
  pages={952--966},
  year={2025}
}

@article{son2023llm,
  title={{LLM}-as-a-judge reward model: What they can and cannot do, 2024},
  author={Son, Guijin and Ko, Hyunwoo and Lee, Hoyoung and Kim, Yewon and Hong, Seunghyeok},
  journal={URL https://arxiv. org/abs/2409.11239},
  year={2023}
}

@article{saha2025learning,
  title={Learning to plan \& reason for evaluation with thinking {LLM}-as-a-judge},
  author={Saha, Swarnadeep and Li, Xian and Ghazvininejad, Marjan and Weston, Jason and Wang, Tianlu},
  journal={arXiv preprint arXiv:2501.18099},
  year={2025}
}

@inproceedings{raju2024constructing,
  title={Constructing domain-specific evaluation sets for {LLM}-as-a-judge},
  author={Raju, Ravi Shanker and Jain, Swayambhoo and Li, Bo and Li, Jonathan Lingjie and Thakker, Urmish},
  booktitle={Proceedings of the 1st Workshop on Customizable NLP: Progress and Challenges in Customizing NLP for a Domain, Application, Group, or Individual (CustomNLP4U)},
  pages={167--181},
  year={2024}
}

@article{zhao2025one,
  title={One token to fool {LLM}-as-a-judge},
  author={Zhao, Yulai and Liu, Haolin and Yu, Dian and Kung, Sunyuan and Chen, Meijia and Mi, Haitao and Yu, Dong},
  journal={arXiv preprint arXiv:2507.08794},
  year={2025}
}

@article{schroeder2024can,
  title={Can you trust {LLM} judgments? reliability of LLM-as-a-judge},
  author={Schroeder, Kayla and Wood-Doughty, Zach},
  journal={arXiv preprint arXiv:2412.12509},
  year={2024}
}

@article{wei2024systematic,
  title={Systematic evaluation of {LLM}-as-a-judge in LLM alignment tasks: Explainable metrics and diverse prompt templates},
  author={Wei, Hui and He, Shenghua and Xia, Tian and Liu, Fei and Wong, Andy and Lin, Jingyang and Han, Mei},
  journal={arXiv preprint arXiv:2408.13006},
  year={2024}
}

@article{yu2025improve,
  title={Improve {LLM}-as-a-judge ability as a general ability},
  author={Yu, Jiachen and Sun, Shaoning and Hu, Xiaohui and Yan, Jiaxu and Yu, Kaidong and Li, Xuelong},
  journal={arXiv preprint arXiv:2502.11689},
  year={2025}
}

@inproceedings{wang2025mllm,
  title={{MLLM}-as-a-judge for image safety without human labeling},
  author={Wang, Zhenting and Hu, Shuming and Zhao, Shiyu and Lin, Xiaowen and Juefei-Xu, Felix and Li, Zhuowei and Han, Ligong and Subramanyam, Harihar and Chen, Li and Chen, Jianfa and others},
  booktitle={Proceedings of the Computer Vision and Pattern Recognition Conference},
  pages={14657--14666},
  year={2025}
}

@inproceedings{chen2024mllm,
  title={{MLLM}-as-a-judge: Assessing multimodal LLM-as-a-judge with vision-language benchmark},
  author={Chen, Dongping and Chen, Ruoxi and Zhang, Shilin and Wang, Yaochen and Liu, Yinuo and Zhou, Huichi and Zhang, Qihui and Wan, Yao and Zhou, Pan and Sun, Lichao},
  booktitle={Forty-first International Conference on Machine Learning},
  year={2024}
}

@article{narayanan2025bias,
  title={Bias in the Picture: Benchmarking VLMs with Social-Cue News Images and LLM-as-Judge Assessment},
  author={Narayanan, Aravind and Khazaie, Vahid Reza and Raza, Shaina},
  journal={arXiv preprint arXiv:2509.19659},
  year={2025},
  doi={}
}

@inproceedings{abdelfattah2020ttpla,
  title={{TTPLA}: An aerial-image dataset for detection and segmentation of transmission towers and power lines},
  author={Abdelfattah, Rabab and Wang, Xiaofeng and Wang, Song},
  booktitle={Proceedings of the Asian conference on computer vision},
  year={2020},
  doi={}
}

@article{abdelfattah2023plgan,
  title={{PLGAN}: Generative adversarial networks for power-line segmentation in aerial images},
  author={Abdelfattah, Rabab and Wang, Xiaofeng and Wang, Song},
  journal={IEEE Transactions on Image Processing},
  volume={32},
  pages={6248--6259},
  year={2023},
  doi={10.1109/TIP.2023.3321465}
}

@inproceedings{hossain2025evaluating,
  title={Evaluating Prompt Engineering for Generalized Power Line Segmentation},
  author={Hossain, Akram and Hasan, Murad and Abdelfattah, Rabab and Scott, Deja and Abdelfatah, Kareem and Sherif, Ahmed},
  booktitle={SoutheastCon 2025},
  pages={508--513},
  year={2025},
  DOI={10.1109/SoutheastCon56624.2025.10971697}
}

@misc{openai_gpt4o,
  author       = {{OpenAI}},
  title        = {{GPT}-4o Model Documentation},
  year         = {2024},
  howpublished = {\url{https://platform.openai.com/docs/models/gpt-4o}},
  note         = {Accessed: Jan. 2026}
}

@article{foundation2026,
  title={From Classical Pipelines to Promptable Foundation Models: A Cross-Domain Survey of Thin-Object
Segmentation for Power Lines, Cracks, and Retinal Vessels},
  author={Hossain, Akram and N. Maharjan and Abdelfattah, Rabab and Ezz-Eldin, Mai and Wang, Xiaofeng and Fouda, Mostafa and  Abdelfatah, Kareem},
  journal={IEEE Internet of Things Journal},
  year={2026}
}
\end{document}